\documentclass[conference]{IEEEtran}
\IEEEoverridecommandlockouts
\usepackage{cite}
\usepackage{amsmath,amssymb,amsfonts}
\usepackage{algorithmic}
\usepackage{graphicx}
\usepackage{textcomp}
\usepackage{xcolor}
\usepackage[a4paper, total={184mm,239mm}]{geometry}

\usepackage{mathtools}
\usepackage{microtype}
\usepackage{booktabs}
\usepackage{enumitem}
\usepackage{tabularx}
\usepackage{hyperref}

\usepackage[ruled,vlined]{algorithm2e} 

\newcommand{\LogHD}{\textsc{LogHD}}

\def\BibTeX{{\rm B\kern-.05em{\sc i\kern-.025em b}\kern-.08em
    T\kern-.1667em\lower.7ex\hbox{E}\kern-.125emX}}
\begin{document}

\title{\textsc{LogHD}: Robust Compression of Hyperdimensional Classifiers via Logarithmic Class-Axis Reduction}

\author{\IEEEauthorblockN{Sanggeon Yun}
\IEEEauthorblockA{\textit{Dept. of Computer Science} \\
\textit{University of California, Irvine}\\
Irvine, CA, USA \\
sanggeoy@uci.edu}
\and
\IEEEauthorblockN{Hyunwoo Oh}
\IEEEauthorblockA{\textit{Dept. of Computer Science} \\
\textit{University of California, Irvine}\\
Irvine, CA, USA \\
hyunwooo@uci.edu}
\and
\IEEEauthorblockN{Ryozo Masukawa}
\IEEEauthorblockA{\textit{Dept. of Computer Science} \\
\textit{University of California, Irvine}\\
Irvine, CA, USA \\
	rmasukaw@uci.edu}
\and
\IEEEauthorblockN{Pietro Mercati}
\IEEEauthorblockA{
\textit{Intel Corporation}\\
Hillsboro, OR, USA \\
pietromercati@gmail.com}
\and
\IEEEauthorblockN{Nathaniel D. Bastian}
\IEEEauthorblockA{\textit{Dept. of Electrical Engineering and Computer Science } \\
\textit{United States Military Academy}\\
West Point, NY, USA \\
nathaniel.bastian@westpoint.edu}
\and
\IEEEauthorblockN{Mohsen Imani}
\IEEEauthorblockA{\textit{Dept. of Computer Science} \\
\textit{University of California, Irvine}\\
Irvine, CA, USA \\
m.imani@uci.edu}
}

\maketitle

\begin{abstract}
    Hyperdimensional computing (HDC) suits memory, energy, and reliability-constrained systems, yet the standard ``one prototype per class'' design requires $\mathcal{O}(CD)$ memory (with $C$ classes and dimensionality $D$). Prior compaction reduces $D$ (feature axis), improving storage/compute but weakening robustness. We introduce \LogHD{}, a logarithmic class-axis reduction that replaces the $C$ per-class prototypes with $n\!\approx\!\lceil\log_k C\rceil$ bundle hypervectors (alphabet size $k$) and decodes in an $n$-dimensional activation space, cutting memory to $\mathcal{O}(D\log_k C)$ while preserving $D$. \LogHD{} uses a capacity-aware codebook and profile-based decoding, and composes with feature-axis sparsification. Across datasets and injected bit flips, \LogHD{} attains competitive accuracy with smaller models and higher resilience at matched memory. Under equal memory, it sustains target accuracy at roughly $2.5$–$3.0\times$ higher bit-flip rates than feature-axis compression; an ASIC instantiation delivers $\mathbf{498\times}$ energy efficiency and $\mathbf{62.6\times}$ speedup over an AMD Ryzen 9 9950X and $\mathbf{24.3\times}$/$\mathbf{6.58\times}$ over an NVIDIA RTX 4090, and is $\mathbf{4.06\times}$ more energy-efficient and $\mathbf{2.19\times}$ faster than a feature-axis HDC ASIC baseline.
\end{abstract}


\section{Introduction}\label{sec:introduction}

Machine-learning architectures operating under tight memory, energy, and reliability budgets increasingly depend on cross-layer co-design and in-/near-memory substrates. Recent surveys and system studies underline both the promise and the sensitivity of memory-centric ML to device non-idealities and limited precision, motivating methods that reduce footprint while preserving robustness~\cite{lammie2025deep, krestinskaya2024neural, tsai2024architecture, burr2023design, frank2023impact}. Within this context, \emph{hyperdimensional computing (HDC)}—a family of \emph{vector–symbolic architectures (VSA)}—is a strong fit: its operations are inherently parallel and memory-centric, and recent cross-layer workflows and macro-level prototypes indicate practical routes to efficient HDC/VSA hardware in embedded and in-/near-memory settings~\cite{nayan2025hydra, du2025cross}.

HDC encodes data as $D$-dimensional hypervectors and learns via light-weight algebraic operations that are parallel, memory-centric, and empirically tolerant to device noise~\cite{kanerva2009hyperdimensional, yun2024neurohash, imani2019framework, zou2021scalable, wang2023disthd, yun2025hyperdimensional, yun2023hyperdimensional, yun2024hypersense, yun2025missionhd}. A conventional HDC classifier stores one prototype per class and predicts by comparing a query $\phi(\mathbf{x})$ to all $C$ prototypes~\cite{hernandez2021onlinehd}. The resulting memory cost $\mathcal{O}(CD)$ dominates in multi-class, resource-constrained regimes, so shrinking this footprint without eroding accuracy or robustness is central.

\begin{figure}[t]
  \centering
  \includegraphics[width=1.\linewidth]{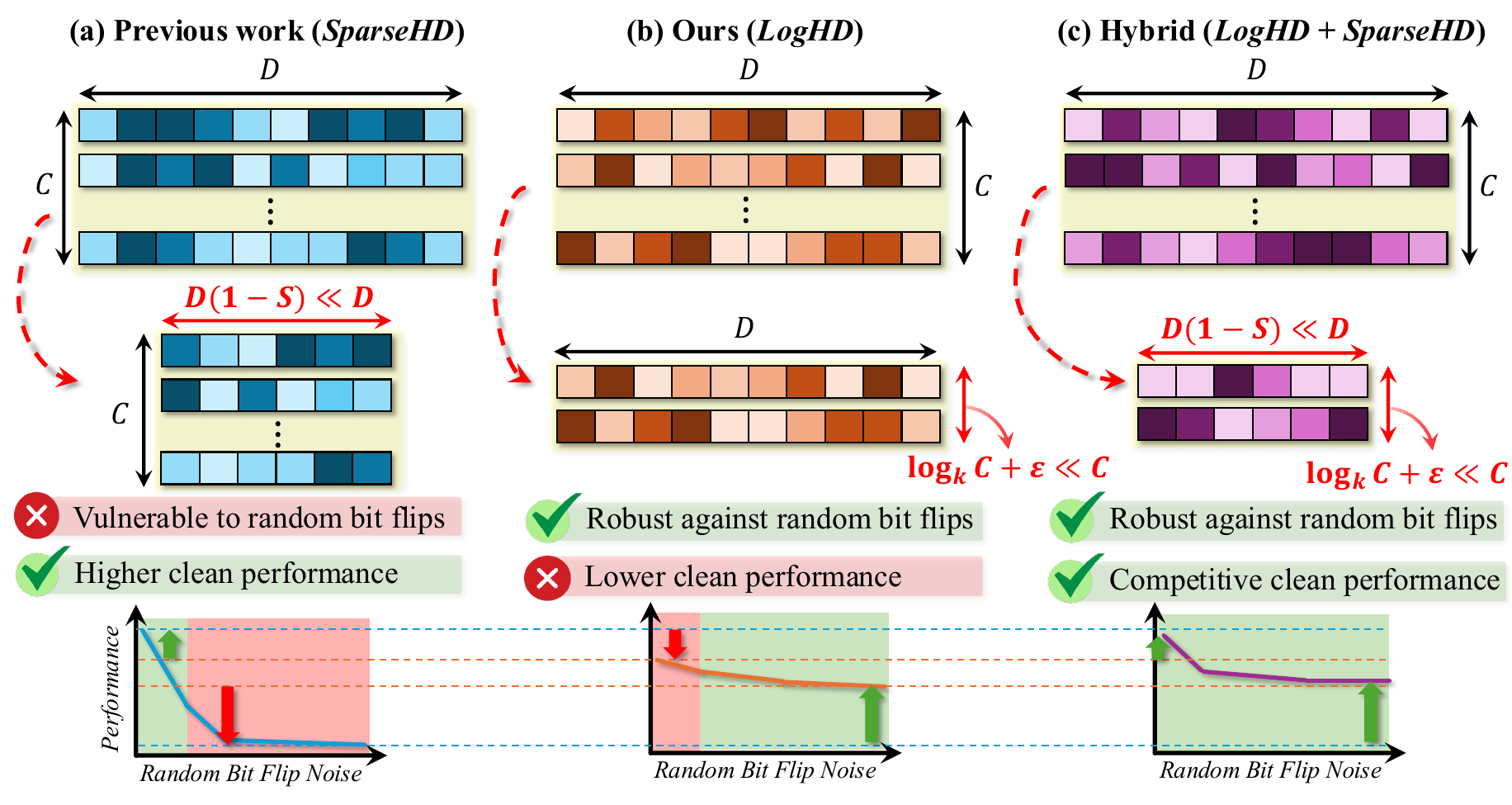}
    \caption{\textbf{Comparison of approaches.} ``Clean performance'' denotes accuracy and throughput measured in fault-free conditions (no injected bit-flip noise). (a) Feature-axis compression (\emph{SparseHD}): reduce $D$ with sparsity $S$; improves storage/compute, but robustness degrades as $D$ shrinks. (b) \emph{\LogHD{}} (class-axis): keep $D$ and replace $C$ prototypes with $n\!\approx\!\lceil\log_k C\rceil{+}\varepsilon$ bundles ($\varepsilon \geq 0$); memory scales logarithmically in $C$ while maintaining high-$D$ robustness. (c) Hybrid (\LogHD{} + SparseHD): combine class- and feature-axis compression to tune memory, clean performance, and robustness.}
  \label{fig:comparison}
\end{figure}

Prior compression largely targets the \emph{feature axis} by reducing $D$~\cite{imani2019sparsehd, morris2019comphd, ge2020classification, verges2025classification}. As a representative state-of-the-art example, \emph{SparseHD}~\cite{imani2019sparsehd} sparsifies trained hypervectors to accelerate computation while keeping one prototype per class. However, operating at lower effective dimensionality weakens robustness to hardware noise—an effect consistent with observations from memory-centric and analog/near-memory accelerators~\cite{burr2023design, frank2023impact, lammie2025deep, krestinskaya2024neural} as illustrated in \autoref{fig:comparison}.\emph{(a)}.

We propose \LogHD{}, a \emph{class-axis} compression scheme that replaces the $C$ per-class prototypes with $n$ bundle hypervectors $\{\mathbf{M}_j\}_{j=1}^n$, where $n\!\ge\!\lceil\log_k C\rceil$ for an alphabet size $k\!\ge\!2$. Each class receives a length-$n$ $k$-ary code that specifies how its prototype contributes to each bundle. At inference, a query is compared to the $n$ bundles to form an activation vector that is decoded via learned per-class profiles. This reduces memory from $\mathcal{O}(CD)$ to $\mathcal{O}(nD)=\mathcal{O}(D\log_k C)$ \emph{without} shrinking $D$. The approach is orthogonal to feature-axis methods (including SparseHD-style sparsification) and aligns with emerging HDC/VSA macros and cross-layer flows that benefit from storing fewer vectors rather than shorter ones~\cite{nayan2025hydra, du2025cross} as depicted in \autoref{fig:comparison}.\emph{(b)}.

Preserving $D$ has direct implications for robustness. In noisy memories and in-/near-memory fabrics, random bit flips or small analog perturbations distort stored hypervectors and degrade similarity computations~\cite{yun2023hyperdimensional, barkam2023reliable}. Concentration-of-measure effects stabilize similarities at large $D$, whereas reducing $D$ amplifies score variance. By decreasing the \emph{number} of stored hypervectors instead of their \emph{length}, \LogHD{} achieves compactness while retaining high-$D$ tolerance to non-idealities, consistent with reliability analyses of analog and near-memory accelerators~\cite{burr2023design, frank2023impact, tsai2024architecture}. Orthogonal techniques—such as flexible numeric types and dynamic dimensional masking—can be layered with \LogHD{} when tighter budgets are required~\cite{li2025fate, liu2025hyperdyn}. Thus, we also study a \emph{hybrid} design that combines \LogHD{} with SparseHD-style sparsification, offering additional memory savings with robustness between pure \LogHD{} and aggressive feature-axis compression as indicated in \autoref{fig:comparison}.\emph{(c)}.

Our evaluation demonstrates that \LogHD{} delivers both compactness and resilience across platforms. Compared to a conventional, non-reduced HDC model, a \LogHD{} ASIC achieves 498$\times$ higher energy efficiency and 62.6$\times$ faster inference than an AMD Ryzen 9 9950X CPU, and 24.3$\times$ and 6.58$\times$ improvements over an NVIDIA RTX 4090 GPU. Against a strong feature-axis baseline, SparseHD, \LogHD{} on ASIC is 4.06$\times$ more energy-efficient and 2.19$\times$ faster under matched conditions. Robustness evaluations further show that, at equal memory budgets, \LogHD{} sustains target accuracy at roughly 2.5–3.0$\times$ higher bit-flip probabilities than feature-axis compression. Finally, a hybrid configuration combining LogHD with SparseHD yields additional memory reduction while achieving intermediate robustness between pure class-axis and pure feature-axis approaches.

\medskip
\noindent We summarize our contributions:
\begin{enumerate}
  \item \textbf{Log-scale class-axis compression.} We introduce \LogHD{}, which replaces the $C$ per-class prototypes with $n\geq\lceil\log_k C\rceil$ bundle hypervectors. This reduces storage and per-query comparisons from $\mathcal{O}(CD)$ to $\mathcal{O}(nD)$ while preserving dimensionality $D$. For example, with $k=3$ and $C=26$, only $n=3$ bundles are required, yielding $8.7\times$ fewer stored prototypes and reducing comparisons per query from $C$ to $n$ without altering the encoder.
  \item \textbf{Balanced decodability with light supervision.} A capacity-aware codebook and a profile-based decoder, complemented by light supervised refinement, balance bundle utilization and mitigate cross-class interference. This preserves reliable classification under superposition with inference cost scaling in $n$ (much smaller than $C$), and integrates naturally with hardware that benefits from storing fewer vectors.
  \item \textbf{Efficiency and robustness.} Across platforms, \LogHD{} achieves up to 498$\times$ higher energy efficiency and 62.6$\times$ faster inference than conventional HDC, and is up to 4.1$\times$ more efficient and 2.2$\times$ faster than the state-of-the-art feature-axis compressor \emph{SparseHD}. At equal memory budgets, it maintains accuracy under up to 3$\times$ higher bit-flip rates, with a hybrid \LogHD{}+SparseHD design offering further memory reduction and intermediate robustness.
\end{enumerate}

\section{Related Work}\label{sec:related}
\subsection{Foundations of HDC}
Hyperdimensional computing (also known as vector–symbolic architectures) represents symbols and structures as dense, high-dimensional hypervectors and manipulates them via binding, bundling, and permutation~\cite{kanerva2009hyperdimensional}. In classification, training samples for a class are superposed into a prototype hypervector; inference compares a query to all class prototypes using cosine or Hamming similarity~\cite{hernandez2021onlinehd}. The paradigm maps well to in-/near-memory accelerators because its operations are linear and massively parallel, and it exhibits empirical resilience to device non-idealities~\cite{yun2023hyperdimensional, barkam2023reliable, barkam2023hdgim}. Comparative studies further catalog encoding/training choices and robustness trends, situating HDC among compact, hardware-friendly learners~\cite{verges2025classification, zou2021scalable, wang2023disthd}.

\subsection{Model-size reduction in HDC}
Most prior work reduces parameters along the \emph{feature axis} while retaining the one prototype per class layout. QuantHD focuses on precision reduction (binary/ternary encodings) so that Hamming distance replaces cosine similarity, improving memory footprint and compare cost without changing the number of stored prototypes~\cite{imani2019quanthd, verges2025classification}. CompHD packs information within each prototype via structured splitting and binding with positional hypervectors followed by superposition; the query path mirrors these transforms at inference~\cite{morris2019comphd}. This yields a complex feature-axis reduction that preserves per-class prototypes and introduces additional binding/unbinding overhead and cross-talk considerations~\cite{verges2025classification}. SparseHD sparsifies trained prototypes and co-optimizes the algorithm with reconfigurable hardware, achieving state-of-the-art efficiency–accuracy trade-offs among few-parameter compression methods while still storing one (sparsified) prototype per class~\cite{imani2019sparsehd, verges2025classification}. In our study, SparseHD is therefore used as the representative feature-axis compressor—both as a strong baseline for performance/robustness comparisons and as the feature-axis component in our hybrid \LogHD{}+SparseHD design—whereas CompHD is not used for hybridization due to its more complex binding/unbinding pipeline and sensitivity to cross-talk. Beyond these, accuracy-oriented updates such as OnlineHD and distributed/regenerative variants improve learning dynamics but leave the parameter layout unchanged, i.e., they do not reduce the number of class prototypes~\cite{hernandez2021onlinehd, zou2021scalable, wang2023disthd, verges2025classification}. In contrast, our approach compresses along the \emph{class axis}: rather than shrinking each prototype, it replaces the $C$ prototypes by $n\!\approx\!\lceil\log_k C\rceil$ bundles and decodes in the induced activation space. To the best of our knowledge and as reflected in recent surveys~\cite{verges2025classification}, log-scale reduction in the \emph{number} of stored class hypervectors has not been explored in HDC.

\subsection{Robustness to hardware noise}
HDC's robustness is rooted in concentration at high dimensionality: for a given bit-flip or analog error rate, larger $D$ averages perturbations and stabilizes similarity scores, whereas reducing $D$ increases score variance~\cite{yun2023hyperdimensional, barkam2023reliable}. Feature-axis compression (CompHD, SparseHD) therefore trades some of this averaging for footprint. By compressing along the class axis and preserving $D$, our method maintains the dimensionality-driven tolerance to non-idealities while still reducing overall model size, a behavior consistent with reliability observations reported for memory-centric ML on in-/near-memory substrates~\cite{verges2025classification}.

\begin{figure*}[t]
    \centering
    \includegraphics[width=0.75\textwidth]{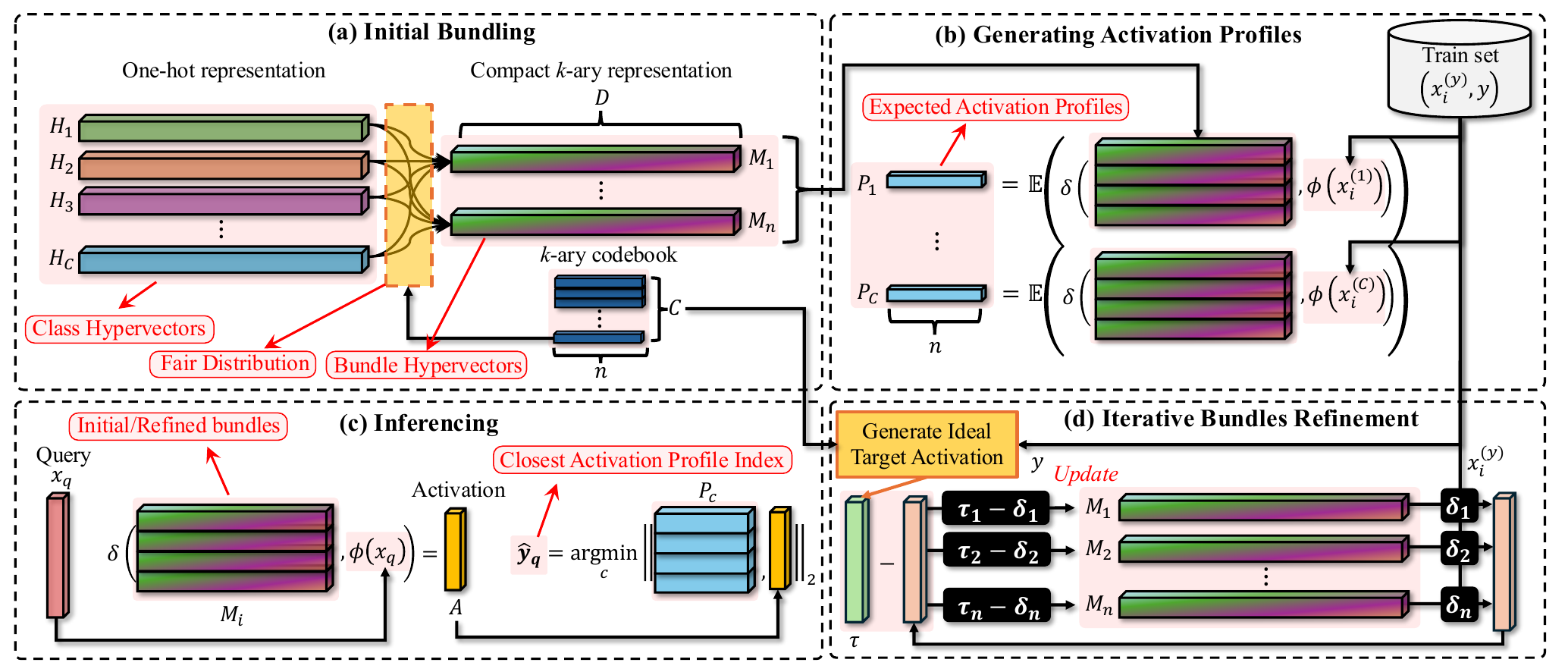}
    \caption{\textbf{\LogHD{} overview.}
  (a) \emph{Initial bundling:} assign each class a length-\(n\) \(k\)-ary code \(B_i\); map symbols via \(g\) and form bundles \(\{\mathbf{M}_j\}\) by weighted superposition; a minimax-load selector balances per-bundle capacity.
  (b) \emph{Activation profiling:} compute activation vectors \(A(\mathbf{x})\) against bundles and estimate per-class means \(\mathbf{P}_y\).
  (c) \emph{Inference:} classify by nearest profile in activation space.
  (d) \emph{Iterative refinement:} nudge bundles toward code-implied targets with a perceptron-style update.
  Replacing \(C\) prototypes by \(n\!\approx\!\lceil\log_k C\rceil\) bundles reduces memory to \(\mathcal{O}(D\log_k C)\) while preserving dimensionality and robustness.}
  \label{fig:overview}
\end{figure*}

\begin{algorithm}[t]
\small
\caption{\LogHD{}: Training and Inference}
\label{alg:loghd}
\DontPrintSemicolon
\SetKwInOut{Input}{Input}\SetKwInOut{Output}{Output}
\Input{Training set $\mathcal{D}=\{(\mathbf{x},y)\}$, encoder $\phi$, alphabet size $k$, \# bundles $n\!\ge\!\lceil\log_k C\rceil$, symbol weight $g(s)=\tfrac{s}{k-1}$, capacity surrogate $U(w)=w^\alpha$, epochs $T$, step size $\eta$.}
\Output{Bundles $\{\mathbf{M}_j\}_{j=1}^{n}$, activation profiles $\{\mathbf{P}_c\}_{c=1}^{C}$, codebook $B\in\{0,\dots,k{-}1\}^{C\times n}$.}

\SetKwBlock{Prot}{(1) Class prototypes}{}
\Prot{$\mathbf{H}_c \leftarrow \sum_{(\mathbf{x},y)\in\mathcal{D},\,y=c}\phi(\mathbf{x})$; \; $\mathbf{H}_c \leftarrow \mathbf{H}_c/\|\mathbf{H}_c\|_2$ for $c=1..C$. }\BlankLine

\SetKwBlock{Code}{(2) Codebook (minimax-load greedy)}{}
\Code{Initialize loads $L_j\leftarrow 0$ for $j=1..n$; candidate pool $\mathcal{Q}\subseteq\{0,\dots,k{-}1\}^n$. \\
\For{$c=1$ \KwTo $C$}{
  $B_{c,:}\leftarrow \arg\min_{s\in \mathcal{Q}} \big[\max_j (L_j + U(g(s_j))) + \varepsilon\,\xi\big]$, \; $\xi\sim\text{Unif}[0,1]$; \\
  $L_j \leftarrow L_j + U(g(B_{c,j}))$ for $j=1..n$; \;
  $\mathcal{Q}\leftarrow \mathcal{Q}\setminus\{B_{c,:}\}$.
} }\BlankLine

\SetKwBlock{Bundle}{(3) Initial Bundling}{}
\Bundle{$\mathbf{M}_j \leftarrow \sum_{c=1}^{C} g(B_{c,j})\,\mathbf{H}_c$; \; $\mathbf{M}_j \leftarrow \mathbf{M}_j/\|\mathbf{M}_j\|_2$ for $j=1..n$. }\BlankLine

\SetKwBlock{Prof}{(4) Activation profiles}{}
\Prof{
For each class $c$: \;$\mathbf{P}_c \leftarrow \frac{1}{|\mathcal{D}_c|}\sum_{(\mathbf{x},y)\in \mathcal{D}_c}\big(\delta(\mathbf{M}_1,\phi(\mathbf{x})),\dots,\delta(\mathbf{M}_n,\phi(\mathbf{x}))\big)$.} \BlankLine

\SetKwBlock{Refine}{(5) Optional refinement (T epochs)}{}
\Refine{
\For{$t=1$ \KwTo $T$}{
  \ForEach{$(\mathbf{x},y)\in\mathcal{D}$}{
    \For{$j=1$ \KwTo $n$}{
      $A_j \leftarrow \delta(\mathbf{M}_j,\phi(\mathbf{x}))$; \;
      $\tau_j \leftarrow 2\,\tfrac{B_{y,j}}{k-1}-1$; \;
      $\mathbf{M}_j \leftarrow \mathbf{M}_j + \eta\,(\tau_j - A_j)\,\phi(\mathbf{x})$; \;
      $\mathbf{M}_j \leftarrow \mathbf{M}_j/\|\mathbf{M}_j\|_2$.
    }
  }
}} \BlankLine

\SetKwBlock{Infer}{(6) Inference $\mathrm{Predict}(\mathbf{x}_q)$}{}
\Infer{
$\mathbf{A}\leftarrow(\delta(\mathbf{M}_1,\phi(\mathbf{x}_q)),\dots,\delta(\mathbf{M}_n,\phi(\mathbf{x}_q)))$; \;
    $\hat{y}\leftarrow \arg\min_{c} \|\mathbf{A}-\mathbf{P}_c\|_2^2$.
}
\end{algorithm}

\section{Methodology}
\label{sec:method}

\subsection{Preliminaries}
\label{subsec:prelim}

Let \(D\) denote the hypervector dimension and \(C\) the number of classes. Conventional HDC stores one prototype (class hypervector) per class, \(\{\mathbf{H}_i \in \mathbb{R}^D\}_{i=1}^{C}\), obtained by superposing encoded training examples of that class. Given a query \(\phi(\mathbf{x})=\mathbf{h}\in\mathbb{R}^D\), the classifier computes scores \(s_i=\delta(\mathbf{h},\mathbf{H}_i)\) and predicts \(\arg\max_i s_i\). Throughout, \(\delta\) is cosine similarity,
\begin{equation}
\delta(\mathbf{u},\mathbf{v})=\Big\langle \tfrac{\mathbf{u}}{\|\mathbf{u}\|_2},\, \tfrac{\mathbf{v}}{\|\mathbf{v}\|_2}\Big\rangle,
\end{equation}
and superposition is the elementwise sum. The memory footprint of this baseline is \(\mathcal{O}(CD)\).

\LogHD{} replaces the set of \(C\) prototypes with \(n\) \emph{bundle hypervectors} \(\{\mathbf{M}_j\}_{j=1}^{n}\), where \(n \ge \lceil \log_k C\rceil\) for a user-chosen alphabet size \(k\ge2\). Each class is assigned a unique length-\(n\) \(k\)-ary code that prescribes how strongly the class prototype contributes to each bundle. A query is compared against the bundles to produce an \(n\)-dimensional activation vector, which is then decoded to a class label. This reduces model memory to \(\mathcal{O}(nD)=\mathcal{O}(D\log_k C)\) without altering the encoder \(\phi\).

\subsection{Overall Pipeline}
\label{subsec:overview}

The complete pipeline (illustrated in \autoref{fig:overview}, panels \emph{a}--\emph{d}) proceeds as follows in concept. First, an initial bundling stage assigns each class a unique \(k\)-ary code and constructs \(n\) bundle hypervectors by weighted superposition of the class prototypes according to the code symbols; this step also balances the per-bundle load induced by the codebook to avoid over-capacity bundles (\autoref{fig:overview}.\emph{(a)}). Second, using the training set, the method estimates a per-class expected activation profile---the mean vector of similarities between encoded examples of that class and the bundles (\autoref{fig:overview}.\emph{(b)}). Third, inference compares a query’s activation vector to these profiles and returns the nearest profile in activation space (\autoref{fig:overview}.\emph{(c)}). Finally, an optional refinement stage performs a small number of supervised updates to the bundles so that activations move toward code-implied targets, thereby mitigating cross-class interference introduced by superposition (\autoref{fig:overview}.\emph{(d)}). Detailed procedures for each stage is described in \autoref{alg:loghd}.

\subsection{Initial Bundling}
\label{subsec:bundling}

\noindent\textbf{Codebook and uniqueness.}
Let \(B\in\{0,1,\dots,k-1\}^{C\times n}\) be a \emph{codebook} whose \(i\)-th row \(B_i=(B_{i,1},\dots,B_{i,n})\) is the unique \(k\)-ary code assigned to class \(i\). Uniqueness requires \(B_i\neq B_{i'}\) for all \(i\neq i'\). When \(C=k^n\), all length-\(n\) codes are used and the codebook is effectively fixed. When \(C<k^n\), there exist \(\binom{k^n}{C}\) valid choices; in this regime the structure of \(B\) strongly influences how many and how strong contributions land on each bundle.

\medskip\noindent\textbf{Load-aware fair code selection.}
To guard against pathological codebooks that overburden a few bundles, \LogHD{} employs a capacity-aware selection heuristic that approximately balances the induced per-bundle load. Define a nonnegative symbol weight \(g:\{0,\dots,k-1\}\!\to\!\mathbb{R}_{\ge0}\) and a nondecreasing capacity surrogate \(U:\mathbb{R}_{\ge0}\!\to\!\mathbb{R}_{\ge0}\). We use \(g(s)=\tfrac{s}{k-1}\) to map symbols to relative contribution strengths and \(U(w)=w^\alpha\) with \(\alpha>0\) to modulate how strongly heavy symbols are penalized in the load objective. For a candidate code \(s=(s_1,\dots,s_n)\), its per-bundle capacity contributions are \(U(g(s_j))\), and the cumulative load on bundle \(j\) induced by the current set of assigned codes \(\mathcal{S}\) is \(L_j=\sum_{c\in\mathcal{S}}U\!\big(g(B_{c,j})\big)\).

The codebook is constructed greedily by repeatedly selecting the next code \(s^\star\) that minimizes the worst-case updated load,
\begin{equation}
s^\star \;=\; \arg\min_{s}\;\max_{1\le j\le n}\Big(L_j + U\big(g(s_j)\big)\Big) \;+\; \varepsilon\,\xi,
\end{equation}
where \(\xi\sim\mathrm{Unif}[0,1]\) provides tie-breaking and diversity, and \(\varepsilon>0\) is a tiny constant. After assigning \(s^\star\) to the next class, we update \(L_j\leftarrow L_j + U(g(s^\star_j))\). When \(k^n\) is moderate we consider the full candidate set; when \(k^n\) is large we draw a sizable random candidate pool, which empirically suffices to flatten the loads while keeping selection time modest. This minimax-load criterion is a direct relaxation of the fair-distribution objective
\begin{equation}
B^\star \;=\; \arg\min_{B} \;\max_{1\le j\le n}\;\sum_{c=1}^{C} U\!\big(g(B_{c,j})\big),
\end{equation}
and is the mechanism by which \LogHD{} avoids over-capacity bundles.

\medskip\noindent\textbf{Constructing the bundles.}
Given the resulting codebook \(B\) and the class prototypes \(\{\mathbf{H}_i\}\), the \(j\)-th bundle is formed by weighted superposition,
\begin{equation}
\label{eq:bundle}
\mathbf{M}_j \;=\; \sum_{i=1}^{C} g\!\big(B_{i,j}\big)\,\mathbf{H}_i,\qquad j=1,\dots,n,
\end{equation}
optionally followed by normalization \(\mathbf{M}_j \leftarrow \mathbf{M}_j / \|\mathbf{M}_j\|_2\) to stabilize cosine similarity. Intuitively, symbol \(0\) contributes nothing to a bundle, whereas larger symbols contribute proportionally more.

\subsection{Generating Activation Profiles}
\label{subsec:profiles}

Because each bundle aggregates multiple classes with heterogeneous weights, single-max decoding is no longer optimal. Instead, \LogHD{} estimates a per-class \emph{expected activation profile}. For a sample \(\mathbf{x}\), define its activation vector against the bundles as
\begin{equation}
A(\mathbf{x}) \;=\; \big(\delta(\mathbf{M}_1,\phi(\mathbf{x})),\,\dots,\,\delta(\mathbf{M}_n,\phi(\mathbf{x}))\big)\in\mathbb{R}^{n}.
\end{equation}
For class \(y\), the expected profile is the conditional mean
\begin{equation}
\mathbf{P}_y \;=\; \mathbb{E}_{\mathbf{x}\mid y}\big[A(\mathbf{x})\big]
\;\approx\; \frac{1}{N_y}\sum_{i=1}^{N_y} A\!\big(\mathbf{x}^{(y)}_i\big),
\end{equation}
where \(\{\mathbf{x}^{(y)}_i\}_{i=1}^{N_y}\) are the training examples of class \(y\).

\subsection{Inference}
\label{subsec:inference}

At test time, a query \(\mathbf{x}_q\) is encoded to \(\phi(\mathbf{x}_q)\), compared with the bundles to obtain \(\mathbf{A}=A(\mathbf{x}_q)\), and decoded by nearest-profile classification in activation space. We adopt the Euclidean metric,
\begin{align}
\begin{split}
\hat{y}_q &\;=\; {\arg\min}_{c\in\{1,\dots,C\}} \big\|\mathbf{A}-\mathbf{P}_c\big\|_2^2
\\&\;=\; \arg\min_{c}\sum_{j=1}^{n}\big(\delta(\mathbf{M}_j,\phi(\mathbf{x}_q))-\mathbf{P}_{c,j}\big)^2,
\end{split}
\end{align}
\label{eq:decode}
which we found to be robust across datasets. Alternatives such as cosine distance in the activation space perform similarly, and a Mahalanobis metric can further help when \(n\) is large relative to \(C\); these variations do not change the training procedure.

\begin{figure*}[t]
    \centering
    \includegraphics[width=0.8\linewidth]{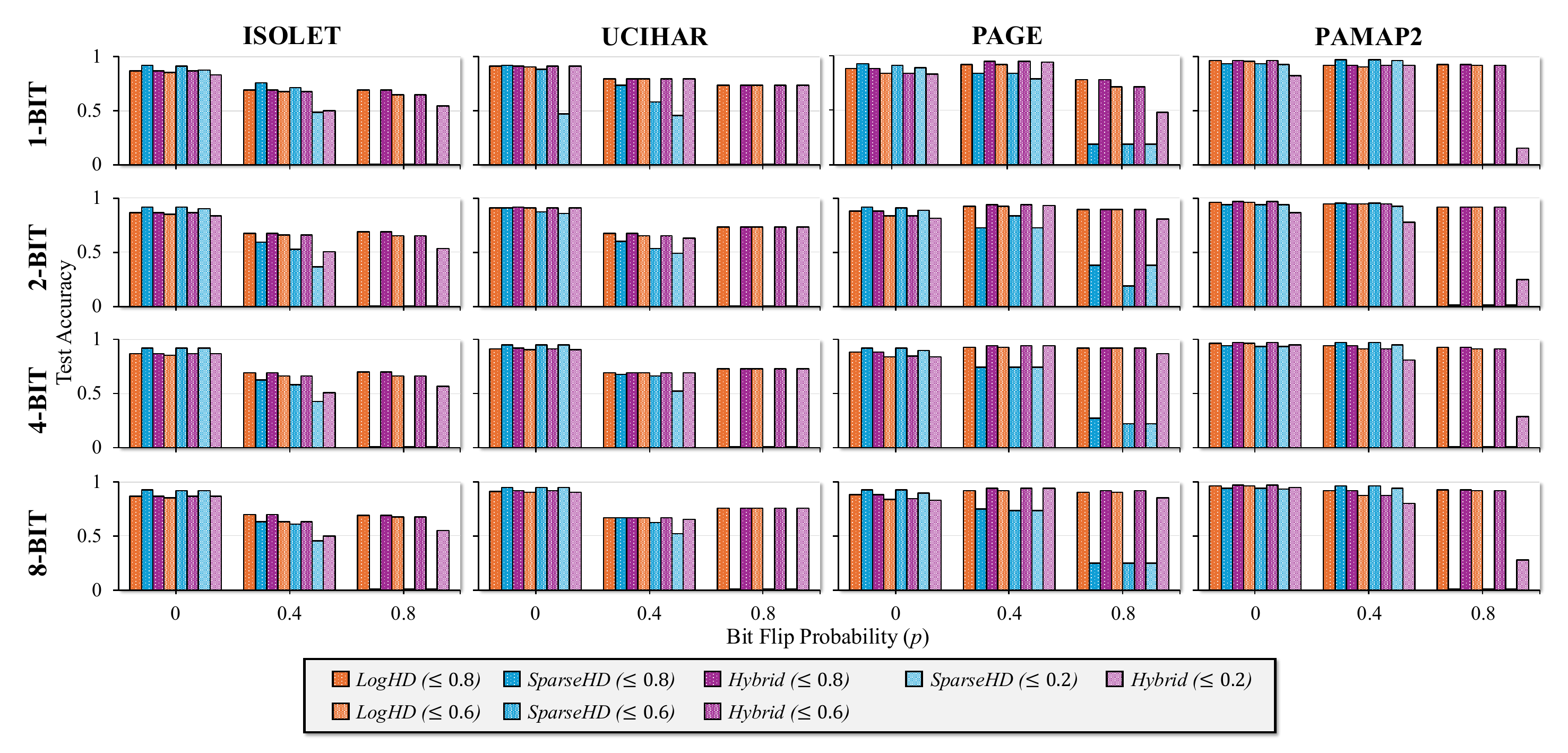}
    \caption{\textbf{Accuracy under random bit flips.} Test accuracy versus bit-flip probability $p$ at matched model-size budgets $(\leq x)$ across datasets, comparing SparseHD, \LogHD{} ($k\in\{2,3\}$), and Hybrid.}
    \label{fig:loghd_datasets}
\end{figure*}

\subsection{Iterative Bundle Refinement}
\label{subsec:refinement}

Initial bundling by \autoref{eq:bundle} may introduce cross-class interference because multiple prototypes are superposed. \LogHD{} therefore supports a light-weight, supervised refinement that nudges each bundle so that the observed activations move toward class-dependent targets implied by the codebook. Let \(t:\{0,\dots,k-1\}\to[-1,1]\) map symbols to target similarities via the linear scaling
\begin{equation}
t(s) \;=\; 2\,\frac{s}{k-1}-1,
\end{equation}
so that \(t(0)=-1\) and \(t(k-1)=1\). For class \(y\), define the target activation vector \(\boldsymbol{\tau}^{(y)} \in \mathbb{R}^{n}\) by \(\tau^{(y)}_j = t\!\big(B_{y,j}\big)\). For each training example \(\mathbf{x}^{(y)}\) we compute activations \(A_j=\delta(\mathbf{M}_j,\phi(\mathbf{x}^{(y)}))\) and update each bundle with a perceptron-style correction,
\begin{equation}
\label{eq:update}
\mathbf{M}_j \;\leftarrow\; \mathbf{M}_j \;+\; \eta\,\big(\tau^{(y)}_j - A_j\big)\,\phi(\mathbf{x}^{(y)}), \qquad j=1,\dots,n,
\end{equation}
optionally followed by normalization \(\mathbf{M}_j \leftarrow \mathbf{M}_j / \|\mathbf{M}_j\|_2\). Here \(\eta>0\) is a learning rate. The term \((\tau^{(y)}_j - A_j)\) increases the projection of \(\mathbf{M}_j\) onto \(\phi(\mathbf{x}^{(y)})\) when the observed activation is below the target and decreases it otherwise. In practice a small number of passes over the training set suffices; excessive refinement may overfit the activation profiles.

\subsection{Complexity and Memory}
\label{subsec:complexity}

Conventional HDC requires \(\mathcal{O}(CD)\) memory for the prototypes and, per query, \(C\) similarities of \(D\)-dimensional vectors. \LogHD{} stores \(n\) bundles and therefore uses \(\mathcal{O}(nD)\) memory with \(n=\lceil \log_k C\rceil + \varepsilon\), where a tiny redundancy \(\varepsilon\in\{0,1,2\}\) is sometimes added for robustness at negligible cost. Constructing the bundles by \autoref{eq:bundle} costs \(\mathcal{O}(nCD)\) arithmetic operations, dominated by superposition. The fair code selection considers \(n\) positions per candidate; with a candidate pool of size \(|\mathcal{Q}|\) the selection pass is \(\mathcal{O}(|\mathcal{Q}|\,n + Cn)\), and \(|\mathcal{Q}|=k^n\) is feasible for moderate \(n\) while random subsampling is effective for larger \(k^n\). Profile estimation requires encoding the training set and computing \(n\) similarities per example, which is \(\mathcal{O}(nND)\) where \(N\) is the number of training examples; accumulation of the class-wise means is negligible by comparison. Inference computes \(n\) similarities and then \(C\) distances in \(\mathbb{R}^{n}\). Because $n \ll C$ in the regimes of interest, \LogHD{} reduces both memory footprint and query-time computation, scaling logarithmically with $C$ rather than linearly as in conventional HDC.

\subsection{Practical Considerations}
\label{subsec:practical}

Normalization substantially improves stability when using cosine similarity; we normalize \(\phi(\mathbf{x})\), \(\mathbf{H}_i\), and \(\mathbf{M}_j\) after construction and after each refinement update. The choices \(g(s)=\tfrac{s}{k-1}\) and \(U(w)=w^\alpha\) with \(\alpha\in[1,2]\) work well empirically; larger \(\alpha\) penalizes heavy symbols more aggressively during code selection and promotes flatter bundle loads. Adding one or two redundant bundles beyond \(\lceil \log_k C\rceil\) often yields a small but reliable accuracy gain by improving separability in activation space without materially affecting the memory advantage.

\begin{figure*}[t]
    \centering
    \includegraphics[width=0.8\linewidth]{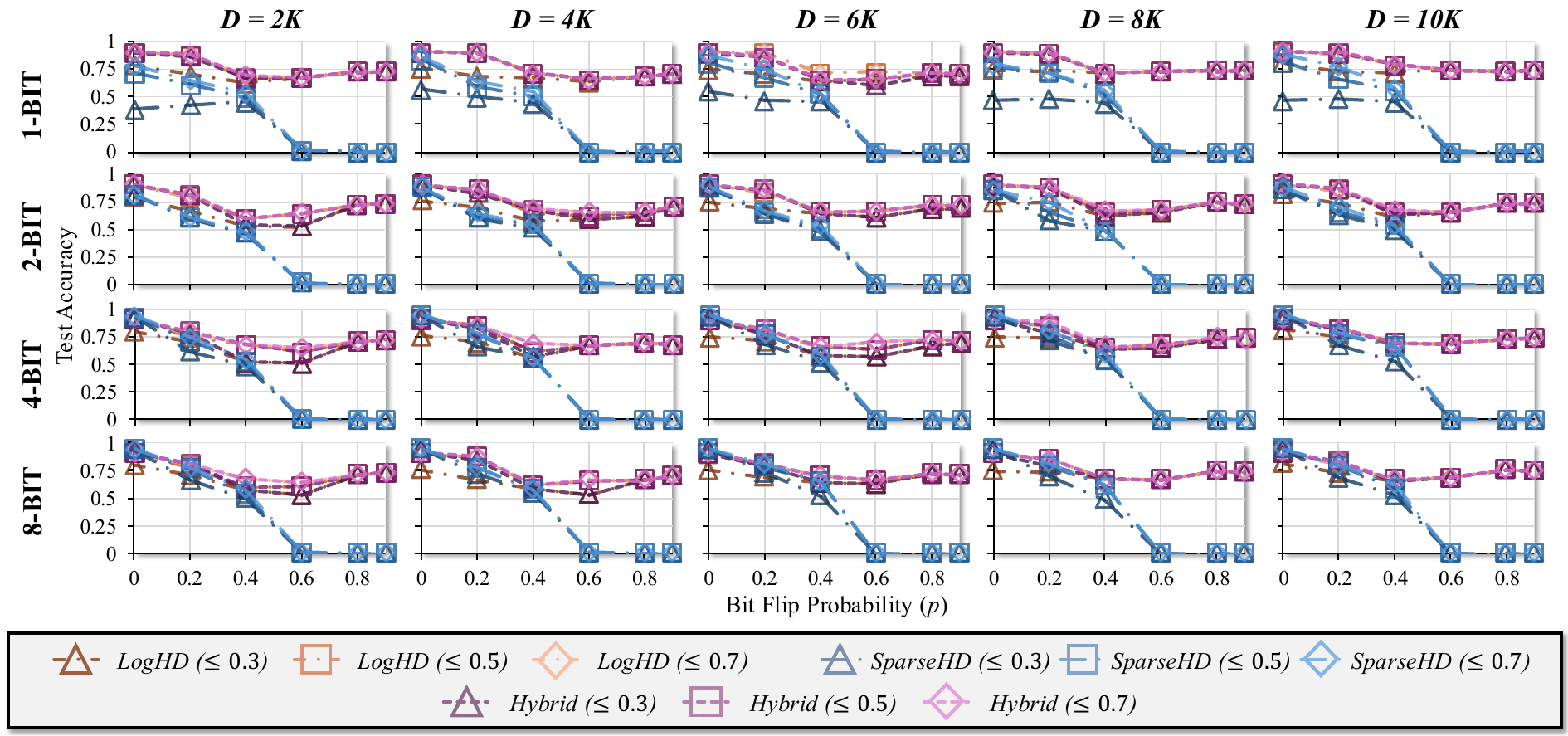}
    \caption{\textbf{Sensitivity to dimensionality and quantization.} Test accuracy on UCIHAR versus bit-flip probability $p$ for varying hypervector dimensionality $D$ and numeric precision (1, 2, 4, 8 bits) at matched model-size budgets $(\leq x)$.}
    \label{fig:loghd_dim}
\end{figure*}

\begin{table}[t]
\centering
\caption{Datasets used in evaluations. $C$ denotes the number of classes.}
\label{tab:datasets}
\resizebox{\columnwidth}{!}{
\begin{tabular}{lrrrrl}
\toprule
\textbf{Dataset} & \textbf{\# Features} & $\boldsymbol{C}$ & \textbf{\# Train} & \textbf{\# Test} & \textbf{Description} \\
\midrule
ISOLET~\cite{isolet_54}  & 617 & 26 & 6{,}238   & 1{,}559   & Voice recognition \\
UCIHAR~\cite{human_activity_recognition_using_smartphones_240}  & 261 & 12 & 6{,}213   & 1{,}554   & Activity recognition (mobile) \\
PAMAP2~\cite{pamap2_physical_activity_monitoring_231}  & 75  & 5  & 611{,}142 & 101{,}582 & Activity recognition (IMU) \\
PAGE~\cite{page_blocks_classification_78}    & 10  & 5  & 4{,}925   & 548      & Page layout blocks classification \\
\bottomrule
\end{tabular}
}
\end{table}

\section{Experiments}\label{sec:experiments}

\begin{figure}
    \centering
    \includegraphics[width=1.0\linewidth]{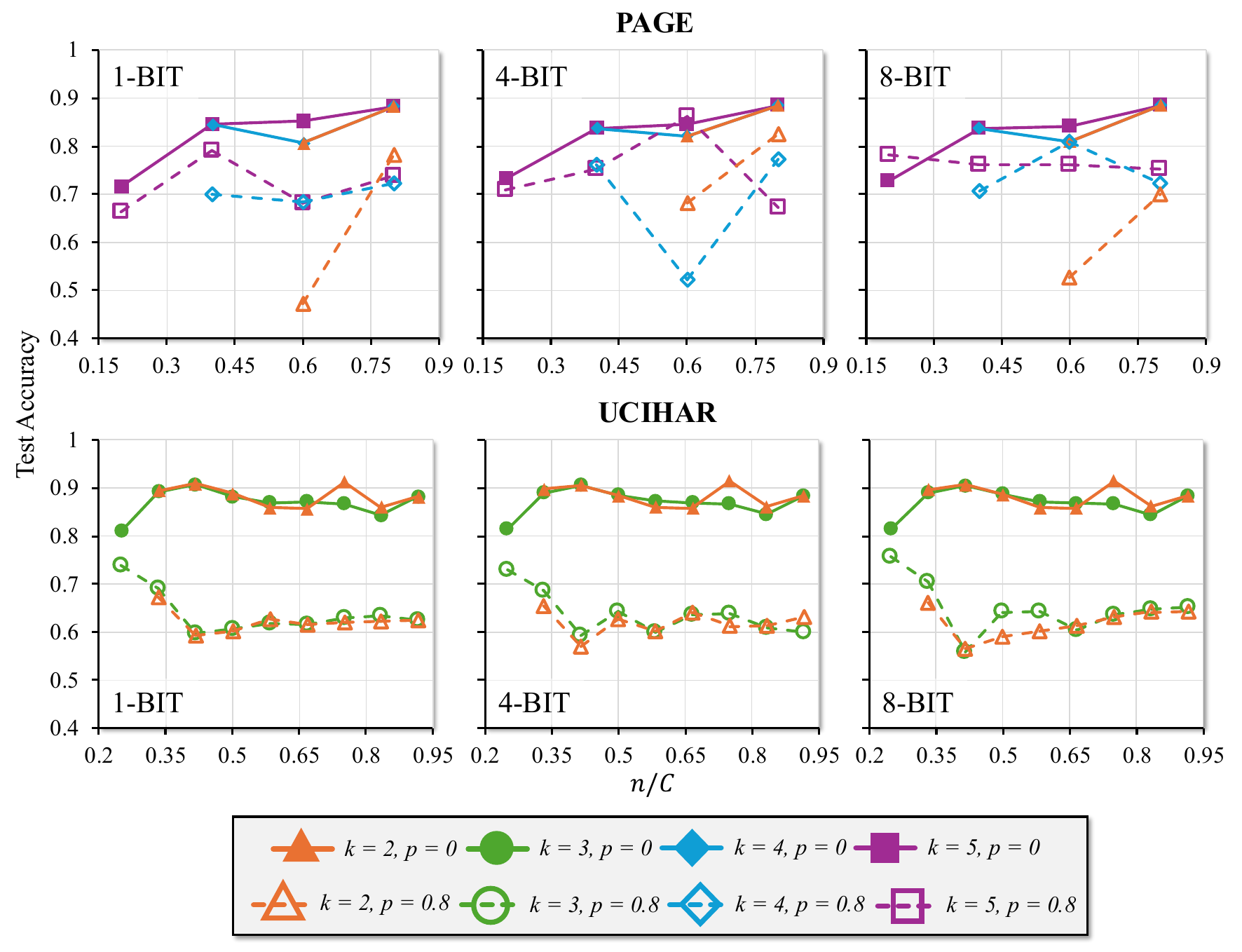}
    \caption{\textbf{Effect of alphabet size $k$.} Test accuracy on PAGE and UCIHAR while varying $n/C$ for different $k$, bit precisions, and flip probabilities $p\in\{0,0.8\}$. For each $k$, the curve sweeps $n$ starting at the feasibility limit $n\ge\lceil\log_k C\rceil$.}
    \label{fig:loghd_k}
\end{figure}

\begin{figure*}[t]
    \centering
    \includegraphics[width=1.0\linewidth]{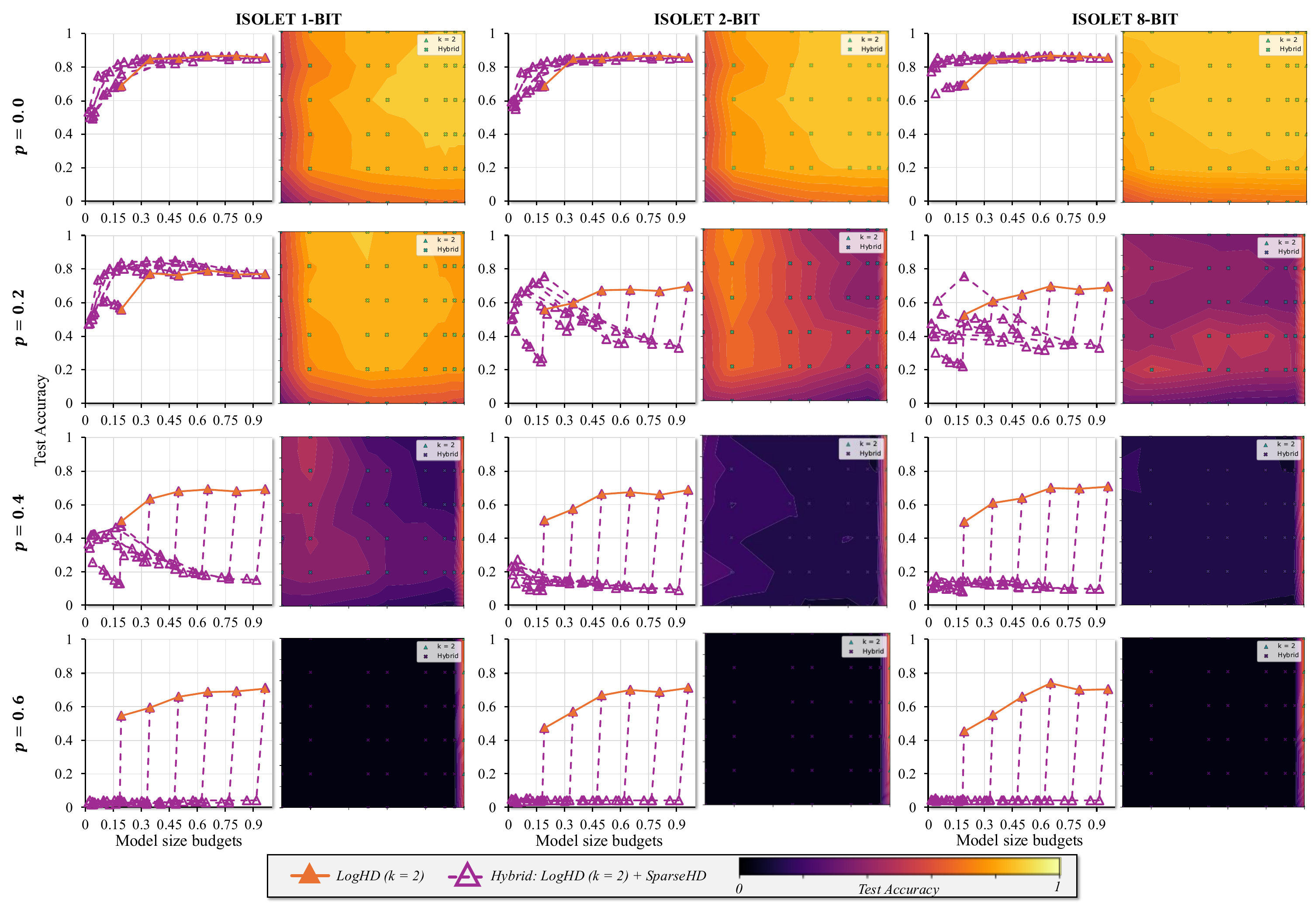}
    \caption{\textbf{Hybrid class- and feature-axis compression.} Test accuracy on ISOLET for \LogHD{} with SparseHD-style sparsification, shown across bit precisions, flip probabilities, number of bundles $n$, and sparsity levels $(1-S)$. The dotted traces showing accuracy as sparsity varies for a fixed base \LogHD{} model $(n,k)$. Heatmaps summarize accuracy as a function of the number of bundles $n$ (rows) and retained feature fraction $(1-S)$ (columns).}
    \label{fig:loghd_hybrid}
\end{figure*}

\subsection{Experimental Setup}

All models are implemented in \texttt{NumPy}. Evaluations use the datasets in \autoref{tab:datasets}. Training is performed in 32-bit floating point; for each target precision (1, 2, 4, 8 bits) we apply post-training quantization to the learned model parameters and then evaluate on the test set. Conventional HDC, \LogHD{}, and SparseHD employ the same encoder and optimization hyperparameters to isolate the effect of the compaction mechanism. Iterative refinement runs for 100 passes over a randomly ordered training set with learning rate $3\times 10^{-4}$ and we set $\alpha=1$ in the capacity surrogate $U$. SparseHD uses dimension-wise sparsification only. Random bit flips are injected into the \emph{stored model state} prior to each test evaluation: for SparseHD the flips are applied to non-pruned coordinates, and for \LogHD{} they are applied to both the bundle hypervectors and the stored activation profiles. Test inputs are not corrupted. Unless noted, we fix $D=10{,}000$ and consider $k\in\{2,3\}$ for \LogHD{}.

\subsection{Performance under Bit-Flip Noise}

\autoref{fig:loghd_datasets} reports test accuracy versus bit-flip probability $p$ at matched model-size budgets across datasets, comparing SparseHD, \LogHD{} ($k\in\{2,3\}$), and a Hybrid composition. Budgets are reported in parentheses as $(\leq x)$, where $x$ is the fraction of the conventional HDC footprint $C\times D$; we fix the dimensionality to $D=10{,}000$. For \LogHD{}—which replaces $C$ prototypes with $n\ge\lceil\log_k C\rceil$ bundles—the minimum feasible budget is $\lceil\log_k C\rceil/C$. With $k\in\{2,3\}$ and datasets having $C{=}5$ (e.g., \textsc{PAGE}, \textsc{PAMAP2}), this lower bound is $2/5=0.4$, explaining the absence of a $(\le 0.2)$ \LogHD{} point unless $k$ is increased. Across all budgets, \LogHD{} sustains accuracy at higher fault rates by preserving $D$ while reducing the number of stored vectors; SparseHD is competitive in the clean setting but degrades rapidly as $p$ increases due to reduced effective dimensionality; the Hybrid lies between these two, extending compaction beyond \LogHD{} with smaller robustness penalties than pure feature-axis compression.

\autoref{fig:loghd_dim} further examines sensitivity to dimensionality and quantization on UCIHAR under fixed memory budgets. We observe a clear trend: higher numeric precision and larger hypervector dimensionality $D$ improve clean accuracy, while higher precision at small $D$ leaves less redundancy per hypervector and thus amplifies the effect of bit flips, yielding larger accuracy drops under noise or dimensionality reduction. Although \LogHD{} can trail slightly in the fault-free setting, it exhibits markedly stronger robustness across all tested $D$, sustaining accuracy even at severe flip probabilities. The Hybrid composition mitigates \LogHD{}'s clean-accuracy gap, delivering competitive and often superior accuracy to SparseHD across dimensionalities and precisions while retaining much of \LogHD{}'s robustness. Taken together, these results indicate that combining class-axis compression---strong against bit-flip noise and quantization---with moderate feature-axis reduction---strong at preserving clean accuracy---yields compact models that are both high-performing and resilient at low memory budgets.

\subsection{Varying the Alphabet Size \texorpdfstring{\boldmath{$k$}}{k}}

\autoref{fig:loghd_k} shows how alphabet size $k$ influences accuracy as the budget scales with $n$. In the fault-free case ($p{=}0$), performance is nearly identical across $k$ once $n \ge \lceil \log_k C \rceil$, suggesting that profile-based decoding already ensures separability and that clean accuracy is largely task-dependent. Under high fault rates ($p{=}0.8$), larger alphabets yield higher accuracy near feasibility. We hypothesize that this stems from two factors: (i) larger $k$ reduces the required code length ($n \approx \lceil \log_k C \rceil$), lowering activation dimensionality and thereby the number of independently corrupted coordinates; and (ii) multi-level coding provides finer weight granularity, enabling the capacity-aware selector to distribute bundle loads more evenly and mitigate cross-class interference. These advantages are most pronounced in noisy, memory-limited regimes (small $n$) and diminish as $n$ increases, where noise accumulation and narrower target steps reduce the margin benefits.

\subsection{Hybrid Class- and Feature-Axis Compression}

\autoref{fig:loghd_hybrid} evaluates combining \LogHD{} with SparseHD-style sparsification. At low fault rates ($p{\le}0.2$), the hybrid can outperform pure \LogHD{}, as pruning weak coordinates regularizes cross-class leakage and sometimes improves clean accuracy. At higher fault rates ($p{\ge}0.4$), it degrades more quickly since reducing the effective dimensionality $(1{-}S)D$ weakens the concentration-of-measure effects that stabilize cosine similarity, underscoring \LogHD{}'s advantage of preserving $D$. Sweeping sparsity at fixed $n$ produces a U-shaped trend: light pruning removes signal and lowers accuracy, moderate pruning suppresses nuisance components and recovers accuracy up to an optimum, and aggressive pruning collapses class separation. The optimum shifts with precision, as lower bit-widths benefit more from sparsification (filtering quantization noise) while higher precisions tolerate denser representations before cross-talk dominates. Overall, reducing along the class axis (smaller $n$) preserves robustness predictably across settings, whereas reducing along the feature axis (larger $S$) introduces sharp uncertainty under faults; the hybrid offers a tunable middle ground for tighter memory budgets, but its robustness ceiling remains bounded by the dimensionality reduction it imposes.

\subsection{Efficiency Gains Across Platforms}

\begin{table}[t]
\centering
\caption{Hardware efficiency ratios of \LogHD{} (ASIC) relative to baselines on ISOLET ($C{=}26$, $k{=}2$). Ratios are \emph{\LogHD{}}/\emph{Baseline}.}
\label{tab:loghd_hw_efficiency}
\resizebox{\columnwidth}{!}{%
\begin{tabular}{lccc}
\toprule
\textbf{Baseline} & \textbf{Platform} & \textbf{Energy eff. ($\times$)} $\uparrow$ & \textbf{Speedup ($\times$)} $\uparrow$ \\
\midrule
SparseHD & ASIC  & 4.06 & 2.19 \\
Conventional HDC & CPU (Ryzen 9 9950X) & 498.1 & 62.6 \\
Conventional HDC & GPU (RTX 4090) & 24.3 & 6.58 \\
\bottomrule
\end{tabular}
}
\end{table}

\autoref{tab:loghd_hw_efficiency} summarizes hardware-level efficiency of \LogHD{} relative to strong baselines. Compared to SparseHD on the same ASIC class, \LogHD{} requires only about one quarter of the energy and half the latency, corresponding to $4.06\times$ higher energy efficiency and $2.19\times$ speedup. Relative to conventional HDC on general-purpose processors, the gains are substantially larger: up to $498\times$ energy efficiency and $63\times$ speedup versus CPU, and $24\times$ and $6.6\times$ versus GPU. These results confirm that compressing along the class axis while preserving dimensionality not only improves robustness under faults but also translates into favorable energy–latency trade-offs when mapped to dedicated hardware.

\section{Conclusions}\label{sec:conclusions}
We presented \LogHD{}, a class-axis compression scheme for hyperdimensional computing that reduces memory and compute from $\mathcal{O}(CD)$ to $\mathcal{O}(D\log_k C)$ while preserving dimensionality. By storing fewer hypervectors rather than shorter ones, \LogHD{} achieves robustness to device faults and quantization, and in hybrid form offers additional flexibility under tight budgets. Experiments across multiple datasets show that \LogHD{} sustains accuracy under up to $3\times$ higher bit-flip rates than feature-axis compression, and ASIC implementations deliver up to $498\times$ energy efficiency and $63\times$ speedup over CPU baselines. Together, these results position class-axis compaction as a scalable, hardware-friendly approach for noise-robust HDC model compression, well suited for TinyML and edge deployments.

\medskip\noindent\textbf{Limitations and Future Work.} This work does not include a formal theoretical analysis of how class-axis reduction interacts with hypervector capacity and separability as dimensionality and class complexity grow. Exploring these relationships, along with richer fault models, is left to future work.

\section*{Acknowledgements}
This work was supported in part by the DARPA Young Faculty Award, the National Science Foundation (NSF) under Grants \#2431561,  \#2127780, \#2319198, \#2321840, \#2312517, and \#2235472, the Semiconductor Research Corporation (SRC), the Office of Naval Research through the Young Investigator Program Award and Grants \#N00014-21-1-2225 and \#N00014-22-1-2067, Army Research Office Grant \#W911NF2410360, and DARPA under Support Agreement No. USMA 23004. Additionally, support was provided by the Air Force Office of Scientific Research under Award \#FA9550-22-1-0253, along with generous gifts from Xilinx and Cisco. 

\bibliographystyle{ieeetr}
\bibliography{mybibliography}

\end{document}